# On Fodor on Darwin on Evolution

Stevan Harnad
Canada Research Chair in Cognitive Sciences
Université du Québec à Montréal
Montréal, Québec, Canada H3C 3P8
http://www.crsc.uqam.ca/
and
Department of Electronics and Computer Science
University of Southampton
Highfield, Southampton, United Kingdom SO17 1BJ
http://www.ecs.soton.ac.uk/~harnad/

**ABSTRACT:** Jerry Fodor argues that Darwin was wrong about "natural selection" because (1) it is only a tautology rather than a scientific law that can support counterfactuals ("If X had happened, Y would have happened") and because (2) only minds can select. Hence Darwin's analogy with "artificial selection" by animal breeders was misleading and evolutionary explanation is nothing but post-hoc historical narrative. I argue that Darwin was right on all counts. Until Darwin's "tautology," it had been believed that either (a) God had created all organisms as they are, or (b) organisms had always been as they are. Darwin revealed instead that (c) organisms have heritable traits that evolved across time through random variation, with survival and reproduction in (changing) environments determining (mindlessly) which variants were successfully transmitted to the next generation. This not only provided the (true) alternative (c), but also the methodology for investigating which traits had been adaptive, how and why; it also led to the discovery of the genetic mechanism of the encoding, variation and evolution of heritable traits. Fodor also draws erroneous conclusions from the analogy between Darwinian evolution and Skinnerian reinforcement learning. Fodor's skepticism about both evolution and learning may be motivated by an overgeneralization of Chomsky's "poverty of the stimulus argument" -- from the origin of Universal Grammar (UG) to the origin of the "concepts" underlying word meaning, which, Fodor thinks, must be "endogenous," rather than evolved or learned.

This is an essay on Jerry Fodor's Hugues Leblanc Lecture Series at UQAM on "What Darwin Got Wrong" (Fodor, forthcoming; Fodor & Piatelli-Palmarini).

I begin with my own 16-point summary of Fodor's position (as I understand it), followed by point-by-point commentary on that summary. (Note that the commentary is on my own version of Fodor's position. I will be happy to correct miconstruals, if any.)

(1) Darwin's theory of evolution is not a "covering-law" scientific theory of the sort we have (for example) with the laws of physics. It does not "support counterfactuals" (i.e., "This is what would have happened if that had happened") in the way Newton's "F = ma" does.

(2) Rather, Darwin's theory looks much closer to a tautology.

(3) Fodor expressed the principle of "Natural Selection" (PNS) in words to the effect that:

> PNS: "There is natural variation of (heritable) traits, and, from that (heritable) variation, '*Natural Selection*' '*selects for*' those traits that confer the greater '*fitness*' (in much the same way that, in Artificial Selection, the animal breeder *selects for* the traits he prefers)."

(4) Fodor's criticism of this principle was that *artificial* selection does indeed work this way: the breeder selects the traits he has in mind, and if they also happen to be correlated with other traits, we can find out exactly which traits he was actually selecting for by asking the breeder which one(s) he was selecting for.

(5) But with natural selection we do not have this, because no one has any traits "in mind." So it makes no sense to say that there was something (namely, increased "fitness") that natural selection somehow "selected for," because "selecting for" is something only minds do, intentionally, and "natural selection" has no mind.

(6) So the theory of natural selection is wrong.

(7) Not only wrong, but empty, since PNS does not predict or explain what will happen in any given situation: We have to look at the actual history in any given case, and then come up with a (possibly true, but post-hoc) explanation of the outcome in that particular case.

(8) And in order to find out which (of potentially very many) correlated traits were the ones that actually caused the organisms that had them to survive and reproduce better, the biologist has to do further experiments and simulations -- something that "natural selection" itself did not do, and has no way of doing.

(9) So evolutionary explanation is really just post-hoc historical explanation; there is no underlying "covering law," and Darwin's notion of "natural selection" is tautological, mentalistic, and explains nothing.

(10) Fodor also pointed out that much the same thing is true (and for much the same reasons) not just of evolution but of "learning" -- in particular, Skinner's reinforcement learning:

(11) When an organism is rewarded by the trainer for doing one thing (choosing a green

triangle) rather than another (choosing something other than a green triangle), it is not even clear -- without further experiments -- what the animal is choosing from among these correlated features (something that's both green and a triangle, or anything that's green, or anything that's triangular, or...), just as it is unclear in evolution which of multiple correlated traits is the adaptive trait.

(12) And people (at least) choose on the basis of what they have in mind (as the animal breeder does).

(13) Skinner was wrong to imagine that the shape that behavior takes is a result of reinforcement, just as Darwin was wrong to think that the shape that organisms take is a result of "natural selection".

(14) Skinner was wrong because reinforcement learning is unable to explain why people do what they do: their mental states (beliefs, desires, etc.) explain it, and Skinner ignored those.

(15) So psychology has the option of studying the true causes of what people do, which are mental (whereas Skinnerian learning explains next to nothing).

(16) But biology does not even have this option of studying the "true" mental causes of evolutionary outcomes, because there are no mental causes, so all that's left is post-hoc historical explanation.

Now some comments:

*(1) Darwin's theory of evolution is not a "covering-law" scientific theory of the sort we have (for example) with the laws of physics. It does not "support counterfactuals" (i.e., "This is what would have happened if that had happened") in the way Newton's "F = ma" does.*

This is true, but I think everyone (including Darwin) already recognized it. The principle of natural selection is not meant to be a "law." Before Darwin, there were two views of why organisms have the traits they have: (i) God created them with those traits (creation theory) or (ii) creatures are as they always were ("steady state" theory).

Darwin suggested the third (and true) alternative which is that, no, creatures did not always have the traits they now have: They evolved that way, in real time, from earlier creatures. Their traits vary from creature to creature, and some of the variants are heritable. So the (heritable) traits of present-day organisms are those that helped their ancestors pass on those very traits more successfully than other (heritable) variants in that ancestral environment:

The methodological consequence of this original and productive insight is that biologists should investigate which traits are heritable, what the mechanism of the heritability is (it turned out to be genes), and what it was in the ancestral environment that made some traits more successfully transmissible than others (and how, and why).

(For some reason, Fodor tends to speak uniformly of “phenotypes” even when he should be saying “genotypes.” An organism’s genotype codes its heritable traits. The organism’s phenotype is the joint result of the expression [through growth and development] of its heritable traits, as modulated by its nonheritable traits, which may be environment-induced or acquired ones. Only heritable [hence genotypic] traits are transmitted genetically, through reproduction, to the next generation – although some phenotypic traits may be transmitted culturally; Sperber & Cladière 2006.)

> *(2) Rather, Darwin's theory looks much closer to a tautology*.

It is not *quite* a tautology, and precisely in the respects that it is *not* a tautology lies its great theoretical and especially methodological value.

That today's heritable traits are those of yesterday's heritable traits that caused the creatures inheriting them to be more successful, in their environment, in surviving and reproducing thanks to (some of) those very traits, is indeed a tautology -- *once you realize the consequences of the fact that there does indeed exist such a heritable variation/retention process;* but not without that realization. And that is the realization we owe to Darwin.

It was neither obvious before Darwin -- nor is it true a-priori, as a matter of necessity -- that there exists heritable variation underlying creatures' traits, and that the survival/reproduction advantages -- i.e., the adaptive advantages -- of those heritable traits in creatures' ancestral environments were what shaped the current traits of creatures across time. Hence it was not obvious that that was what you had to investigate if you wanted to know what traits had evolved in what environments, as a result of what adaptive advantages. But the general principle itself does not identify any particular trait of any particular creature in any particular environment.

> *(3) Fodor expressed the principle of "Natural Selection" (PNS) in words to the effect that*:
>
> > *PNS: "There is natural variation of (heritable) traits, and, from that (heritable) variation, 'Natural Selection' 'selects for' those traits that confer the greater 'fitness' (in much the same way that, in Artificial Selection, the animal breeder selects for the traits he prefers)."*

This formulation of the PNS is correct, but a more explicit and perspicuous way to put it (with more words but fewer metaphors) would be:

> > **PNS: "There is natural variation of (heritable) traits, and, from that (heritable) variation, it is the relative success of the creatures that inherit the traits, *in surviving and reproducing in a particular environment*, that determines which traits are passed on to the next generation (in much the same way that, in Artificial Selection, the animal breeder selects for the traits he prefers)."**

Notice that there has been no need at all to mention either "Natural Selection," which is just a metaphor, nor "fitness," which really just means "the relative success of the heritable variants in a given environment."

Success in survival/reproduction -- determined by *the effects of the environment on the distribution of heritable traits in the next generation* -- is what replaces the intentional choices of the selective animal breeder by a mindless process.

More generally, the case of the effects of the intentional choices of selective breeders is just a very special -- and until very recently, highly atypical -- case of this same, general, mindless process, namely, the transmission success of heritable traits being determined by the causal contingencies of the environment in which they occur: M*indful animal breeding is just one of those environments.* Moreover, *mindfulness itself is just one -- or several -- of those evolved, heritable traits*: It is often the traits of one organism that constitute part of the environment of another organism, whether within or between species.

> *(4) Fodor's criticism of this principle was that* <u>*artificial*</u> *selection does indeed work this way: the breeder selects the traits he has in mind, and if they also happen to be correlated with other traits, we can find out exactly which traits he was actually selecting for by asking the breeder which one(s) he was selecting for.*

> *(5) But with natural selection we do not have this, because no one has any traits "in mind." So it makes no sense to say that there was something (namely, increased "fitness") that natural selection somehow "selected for," because "selecting for" is something minds do, intentionally, and "natural selection" has no mind.*

> *(6) So the theory of natural selection is wrong.*

This is the wrong conclusion to draw.

The Darwinian mechanism of adaptation -- with blind variation of heritable traits in one generation, and then reproductive transmission-success in the given environment determining the distribution of those traits in the next generation -- is an original, ingenious (and true!) explanation not only of how and why creatures have the (heritable) traits they have, but of the relation between evolution in general, and human animal breeding in particular: The distribution of heritable traits is always determined by how successful they make their bearers in surviving and reproducing, but in the special case of human-bred animals it happens to be conscious humans that are "culling out" the creatures with the "maladaptive" traits (rather than, say, hungry predators doing it).

> *(7) Not only wrong, but empty, since PNS does not predict or explain what will happen in any given situation: We have to look at the actual history in any given case, and then come up with a (possibly true, but post-hoc) explanation of the outcome in that particular case.*

It's certainly true that, apart from the original, ingenious (and true!) explanation of how heritable traits evolve in general, Darwin's PNS "merely" provides a methodology for investigating what happened in particular cases (particular traits, particular creatures, particular environments).

> *(8) And in order to find out which (of potentially very many) correlated traits were the ones that actually caused the organisms that had them to survive and reproduce better, the biologist has to do further experiments and simulations – something that "natural selection" itself did not do, and has no way of doing.*

The only thing evolution "does" is change the trait distribution from generation to generation as a function of environmental contingencies and their effects on survival/reproductive success. If there are correlated traits, where one of them is the causally effective one and the others just happen to be coupled with it, evolution could not "care" less (until and unless some of the correlated traits turn into a *disadvantage* for survival and reproduction).

Unlike the calculus of variations or even ordinary engineering, evolution does not optimize; it does not even wield Occam's Razor (at least not in this particular respect): As Herbert Simon (1996) put it, evolution merely "provisionally 'satisfices'" (which just means -- let us hasten to point out, lest someone is again tempted to give it a mentalistic interpretation! -- that the winners, from the prior generation, just have to be better than the current competition, not *all possible* competition; hence no "counterfactual-supportingness" here either! "Fitness" is local, provisional, relative and approximate, not global, permanent, absolute, or exact, like F=ma).

If we are interested in whether it is the fox's swiftness or its (correlated) smooth tail that is the cause of its adaptive success, we will have to manipulate experimentally to find out (as some experimental as well as computational biologists do).

> *(9) So evolutionary explanation is really just post-hoc historical explanation; there is no underlying "covering law," and Darwin's notion of "natural selection" is tautological, mentalistic, and explains nothing.*

Tautological, yes, to a degree. But PNS is not the first tautology to be discovered, and to prove productive, having previously not been known, or noticed. (The law of the excluded middle is perhaps another.) And PNS is not a pure tautology (which would have taken the form: "whatever happens, happens"). There is the restriction to *heritable* traits (rather than all traits), which led, amongst other things, to the discovery of the mechanism of their heritability (genes and DNA).

But, yes, to say that "the distribution of heritable traits in generation N is determined by the inheritance success of the distribution of heritable traits in generation N-1" is tautological; it's just that this (true) tautology had never occurred to anyone before Darwin. They either thought that traits came from God, or that they had always been there and simply varied randomly.

There is nothing mentalistic whatsoever here; insead we have the mechanical key to the explanation of the origin and diversity of all heritable traits.

> *(10) Fodor also pointed out that much the same thing is true (and for much the same reasons) not just of evolution but of "learning" -- in particular, Skinner's reinforcement*

*learning.*

It is true (and has been pointed out by many others too) that both Skinner's reinforcement learning and Darwin's adaptive evolution are based on random-variation plus differential-retention-by-consequences (Catania & Harnad 1988).

But the analogy ends there. Skinner tried to explain the "shape" of *all* behavior as having been determined by random-variation (in responses emitted by the organism) plus reward ("selection by consequences"). Skinner even has an (implicit and unintended) analogy with animal breeders' artificial selection vs "natural selection": the analogy between animal trainers' "artificial reinforcement" (for doing the performance tricks) and the "natural reinforcement" in our (and other animals') natural lives that "rewards and trains us" to perform as we do.

But most natural behavior is not explained or explainable as the result of shaping by Skinnerian reinforcement. In particular, the (universal) grammatical "shape" of language (UG) is not learned or learnable by children through reinforcement training, because of the "poverty-of-the-stimulus": What the child hears and says, and what corrections (reinforcements) it receives during its brief language-learning period, are demonstrably too underdetermined to be able to learn UG from -- either via Skinner reinforcement learning or via any other learning mechanism. There just are not enough data (especially negative data: errors and error-corrections), or time (Chomsky 1980).

The same, however, is not true of the evolution of heritable biological traits: For the evolution of today's organisms' heritable traits (their gene pool) there has been plenty of time, and of positive and negative data, in the 4 billion years since the "primal soup" (Dawkins 1976). In other words, *there is no evolutionary "poverty-of-the-stimulus" problem with the evolution or evolvability of heritable traits, through random variation plus retention as determined by their adaptive consequences.* (My guess is that Fodor, in his heart of hearts, thinks that there *is* such a poverty-of-the-stimulus problem with the evolution of biological traits, and that that is what motivates his scepticism about both learning and evolution! Fodor & Pylyshyn 1988)

Four billion years is plenty more time and data than the child's c. 4 language-learning years for evolving all heritable traits. It is not clear, though, that even that is enough time or data to evolve UG! But that's a very special case. Chomsky (1959) could have refuted Skinner (and did) with a lot more than just that: Reinforcement learning does not explain the "shape" of most nontrivial behavior, probably because a lot of human and animal behavior is not the result of reinforcement learning during the lifetime of the organism. Some of it evolved earlier; and some of human behavior is taught, or self-taught, via reasoning. Besides, even behaviors that are learnable via reinforcement learning are not explained until one provides the internal design of the learner that is capable of learning them (via reinforcement learning): Some behavior may indeed be learnable through trial-and-error-correction alone, but ho*w do we have to be built in order to be capable of learning it* through trial-and-error-correction alone?

This calls for the kind of computational and robotic modeling of our total performance capacity that Turing (1950) called for (and that Skinner's impoverished, a-theoretical "reinforcement schedules" certainly did not and could not provide; Harnad 1996). But that does not mean that there aren't mechanisms for general trial-and-error-correction-based learning. (That is what a lot of computational learning theory, including neural net learning, is all about; Harnad 2009). Since no one yet knows the scope of general learning theory, no one is in a position to say what it can or cannot do, except in special cases where you do have a poverty-of-the-stimulus argument -- which no one does have, either for human learning ability in general, or for heritable traits in general, but only for the single special case of UG.

> *(11) When an organism is rewarded by the trainer for doing one thing (choosing a green triangle) rather than another (choosing something other than a green triangle), it is not even clear -- without further experiments -- what the animal is choosing from among these correlated features (something that's both green and a triangle, or anything that's green, or anything that's triangular, or...), just as it is unclear in evolution which of multiple correlated traits is the adaptive trait.*

This is certainly true. So in the case of learning, you do the further experiments to see which of the correlated features the animal is actually using; and in the case of evolution, you do the further experiments to see which of the correlated traits are actually causing the reproductive/survival advantage.

This is part of doing the actual science. It in no way impugns Darwin with having had to make illicit use of any nonexistent “mindfulness” in any way. It does not even impugn Skinner with that. It is merely another flaw in Skinner that, besides being oblivious to the limits on what can be explained as being just the outcome of reinforcement learning, he ignored the fact that people have minds and often do things because they've made up their minds to do them. Of course, that fact is not an explanation either. But neither is Skinnerian reinforcement an explanation of the nature or origins of our mental powers. (Darwinian evolution, on the other hand, is very likely to be!)

But, in any case, experimentally unbundling and testing correlated variables is not a problem or handicap to either Skinner or Darwin. It is perfectly normal science.

> *(12) And people (at least) choose on the basis of what they have in mind (as the animal breeder does).*

That's bad news for Skinner, because he can't account for our mental capacities with reinforcement learning.

But it's not bad news for Darwin, in any way.

Our cognitive capacities (including our learning capacities -- inasmuch as they are heritable rather than acquired traits) are simply a subset of the traits that evolution (including neural and behavioral evolution) needs to account for. These cognitive capacities are far from having been accounted for yet; but there is nothing at all in Fodor's critique of Darwin (or even of Skinnerian reinforcement learning) that implies that they will not or cannot be accounted for: There is no "poverty-of-the-stimulus" argument here.

And the "intentionality" argument against PNS is not an argument, but a misunderstanding of Darwin's "selection" metaphor.

"Counterfactual-support" for the general PNS is supererogatory. The PNS does not formulate a law that requires counterfactual-support (if anything really does). PNS is simple true, new, and methodologically fruitful (boundlessly fruitful!).

> *(13) Skinner was wrong to imagine that the shape that behavior takes is a result of reinforcement, just as Darwin was wrong to think that the shape that organisms take is a result of "natural selection".*

Skinner was wrong to imagine that the origin and nature of most nontrivial behavior is explicable merely by reinforcement history. Darwin, in contrast, was quite wonderfully right that the origin and nature of most heritable traits could be fully explained by blind variation and retention, based on gentoypes' survival/reproductive success.

> *(14) Skinner was wrong because reinforcement learning is unable to explain why people do what they do: their mental states (beliefs, desires, etc.) explain it, and Skinner ignored those.*

That's part of why Skinner was wrong. But mental states don't explain anything either: they themselves stand in need of explanation. That explanation is almost certainly going to be an adaptive/evolutionary one, based mainly on the *performance capacities* that our evolved mental powers conferred on us, and their contribution to our survival/reproductive success. The only thing that looks inexplicable in this way right now is UG, because of the poverty-of-the-stimulus (Harnad 2008a). Unless UG turns out to be learnable or evolvable, it will remain an unexplained evolutionary anomaly: an inborn trait that was not shaped by adaptive contingencies (Harnad 1976).

[There is one other trait, however, that looks even more likely never to have an adaptive explanation, nor any causal/functional explanation at all, and that is consciousness (i.e., feeling): The fact that our internal functional states -- the ones that have given us the adaptive capacities and advantage that they have given us -- also happen to be conscious states (in other words, *felt* states) is causally inexplicable: Unless feeling turns out to be an independent causal force in the universe -- in other words, unless "telekinetic dualism" turns out to be true (and all evidence to date suggests overwhelmingly that it is not true) -- both the existence and the causal role of feeling are beyond the scope of evolutionary explanation, indeed beyond the scope of any form of causal explanation. Feeling may be one of those "correlated" traits, free-riding on some other effective trait, but it cannot have any independent causal (hence adaptive) power of its own.]

> *(15) So psychology has the option of studying the true causes of what people do, which are mental (whereas Skinnerian learning explains next to nothing).*

Cognitive science and evolution, together, need to explain the origins, nature and underlying computational and neural mechanisms of our performance capacities, both the inherited ones and the learned ones (including our capacity to learn).

Yes, Skinnerian reinforcement explains next to nothing.

> *(16) But biology does not even have this option of studying the "true" mental causes of evolutionary outcomes, because there are no mental causes, so all that's left is post-hoc historical explanation.*

In the case of investigating any particular heritable trait, some specific local research -- historical, functional, ecological, computational -- will no doubt be necessary, plus some experimental hypothesis-testing, as in all areas of science. But that certainly does not make evolutionary explanation "merely" historical: One can do predictive hypothesis-testing; correlated traits can be experimentally disentangled. Analogies and homologies can be and are studied and tested.

This is all characteristic of *reverse engineering* (Dennett 1994; Harnad 1994) In forward engineering, an engineer deliberately designs and builds a device, for a purpose, applying already-known functional (engineering) principles.

In the case of evolution, the device is already there, built, and the objective is to figure out how it works, and how and why it got that way.

No "counterfactual-supporting cover laws" (other than those of physics and chemistry) are needed in either forward engineering or reverse engineering. All they're concerned about is how causal devices work: For their devices, forward engineers already know. For their devices, reverse engineers need to figure it out.

In the case of the human being, we have a causal device that, among other things, has all of our cognitive and behavioral capacities. These include the capacity to recognize, identify, manipulate, name, define, describe and reason about all the objects, events, actions, states and traits that we humans are able to recognize, identify, manipulate, name, define, describe, and reason about (and that includes the natural language capacity to understand and produce all those verbal definitions, descriptions and deductions; Blondin-Massé et al. 2008).

That's a tall order, but for cognitive reverse-engineers it means scaling up, eventually, to a model that can pass the Turing Test -- i.e., can do anything we can do, indistinguishably from any of us (Harnad 2002b).

It's clear that the "Blind Watchmaker" has designed such a device. We are it. But all that is meant by the "Blind Watchmaker" is that mindless mechanism of random variation in heritable traits whose distribution changes from generation to generation as a result of the environment's differential effects on the survival and reproduction of their bearers (Dawkins 1986; Harnad 2002a).

Yes, Skinner made a similar claim about all of behavior being shaped by its consequences, through reinforcement, and was monumentally wrong. But that was because reinforcement alone can be demonstrated to be insufficient to "shape" most nontrivial behavior -- and in the special case of UG, there is even the poverty-of-the-stimulus problem, making learning UG impossible in principle for the child.

Nevertheless, learning (if not reinforcement learning) has nontrivial performance power too, and computational learning theory has already "reverse-engineered" some of it -- enough to allow us to conclude that (in the absence of a poverty-of-the-stimulus problem of which no one has yet provided even a hint -- and neither "vanishing feature intersections" nor "correlated features" prove to be successful stand-ins for this nonexistent poverty-of-the-stimulus problem) the *lexicon is indeed learnable*, with no need for recourse to inborn "whirlpools" in a concept ocean that is "endogenous" to the Big Bang (as Fodor seems to be suggesting).

Ordinary Darwinian evolutionary precursors, in the form of our sensorimotor categorization and manipulation capacities (Harnad 2005), plus the evolution of language (i.e., the evolution of the capacity to string the arbitrary names of our simple categories into truth-valued propositions defining and describing composite categories) are enough to account for the origin of both words and their meanings (Harnad 1976; Cangelosi et al. 2002).

For the origin of species and their heritable traits, PNS continues to be the only viable account on offer. (There are no "endogenous" whirlpools that are place-holders for species and their traits either.)

It's no fun defending a theory that most (non-believers) believe in. But I've had fun defending Darwin against Fodor's critique. It has helped bring Darwin into even clearer focus for me, and I hope it may have the same effect for others.

---

I have received a few responses to the above critique.

---

**Dan Sperber** pointed out that Darwin had written the following on the subject of the intentionality of artificial selection:

> *"At the present time, eminent breeders try by methodical selection, with a distinct object in view, to make a new strain or sub-breed ... But, for our purpose, a form of selection, which may be called unconscious, and which results from everyone trying to possess and breed from the best individual animals, is more important' (Darwin 1872, p. 26)."*

This is a very apt point. The gist of Fodor's's critique is that "natural selection" gives rise to correlated traits, and only further experiment can show which of the correlates were actually causal in enhancing survival and reproduction. Hence "natural selection" itself is not a "counterfactual-supporting" law (indeed, it is not "selection," because selection can only be intentional).

The fact that intentional (artificial) selection is indeed intentional, because we can ask the breeder "what trait did you select for" and he can tell you, truthfully, "I selected for curly tails" does not, of course, rule out either that curly tails are correlated with other traits, traits that the breeder did not realize he was also selecting for; nor does it rule out the slightly more complex case (because it brings out the problematic role of "unconscious intention") that the breeder, in consciously selecting for curly tails, was *also* "unconsciously selecting" for larger curly tails rather than smaller ones. (Nor does it rule out the fact that animal breeders sometimes select for traits without even realizing that they are selecting.)

So if the evolutionary outcome is a shift in the distribution of heritable traits toward trait X, this can be,

(1) in the case of animal breeding:

because (1a) the breeder selected X intentionally and consciously; or because (1b) the breeder selected X unconsciously; or because (1b) the breeder selected some other trait Y intentionally and consciously; but also selected trait X, unconsciously, or because (1d) the breeder selected some other trait Y intentionally and consciously, but trait X happened to be correlated with Y; or because (1e) the breeder selected some other trait Y unconsciously, and trait X happened to be correlated with Y; or

(2) in the case of "natural selection":

because (2a) success in survival and reproduction increased the frequency of trait X; or because (2b) success in survival and reproduction increased the frequency of trait X, but trait Y's frequency also increased, because trait Y was correlated with trait X, and later (human) experiment showed that trait X was the cause of the survival/reproductive success, whereas Y was just a fellow-traveller; or

because (2c) (same as (2b) but switch X and Y)

Clearly, Darwin's own philosophical view on "unconscious selectivity" does not really matter in our discussion of Fodor's critique of Darwin. We have eight subcases, but the point about (1) and (2) is the same: Similar intergenerational changes in heritable trait frequency for X can be caused either (1) by selective breeding on the part of a conscious breeder or (2) by the adaptive consequences, in terms of the survival/reproduction success of the trait itself, in the trait-bearer's natural environment.

In both cases, (1) and (2), there can be correlated traits -- traits (Y) that likewise increase in frequency because they are somehow coupled with trait X -- and only further experiment can show what those traits are, and whether or not they make a causal contribution to the enhanced survival/reproduction success.

In the case of (2), further experiment would determine whether X, Y, both (or neither) caused the increase frequency.

In the case of (1), further (psychological) experiments on the breeders would determine whether they would select as strongly for X if Y were uncoupled from X.

But chances are that Darwin was referring to something more fundamental than the question of conscious vs. unconscious selectivity in human animal breeding, namely that, either way, human selection, whether conscious or unconscious, is itself merely a particular case of "natural selection." This is how I put this point in own commentary:

> Success in survival/reproduction -- determined by *the effects of the environment on the distribution of heritable traits in the next generation* -- is what replaces the intentional choices of the selective animal breeder by a mindless process.
>
> More generally, the case of the effects of the intentional choices of selective breeders is just a very special -- and until very recently, highly atypical -- case of this same, general, mindless process, namely, the transmission success of heritable traits being determined by the causal contingencies of the environment in which they occur: M*indful animal breeding is just one of those environments.* Moreover, *mindfulness itself is just one -- or several -- of those evolved, heritable traits*: It is often the traits of one organism that constitute part of the environment of another organism, whether within or between species.

I might add -- as another complication for the notion that conscious intentions are somehow criterial in any of this: There is both *explicit* learning, in which the human subject learns, and is conscious of, and can verbalize that -- and what -- he has learned, and how; and there is *implicit* learning, in which the subject does indeed learn (in that his performance systematically changes with respect to an external criterion for correctness), but he is not conscious of, and cannot verbalize, that and what he has learned, and how.

Since I am not a believer in unconscious intentionality (and I doubt that William of Occam would have believed in it either -- or should have, if he was an Occamian), I think this suggests that the baggage of intentionality is even more supererogatory here than my explicit analysis has suggested.

But to see this would be to see the truth of something that I so far seem to be quite alone in believing, which is that *the only difference between an intentional system and a system that is systematically interpretable by intentional systems as being an intentional system, but in reality is merely a syntactic system with no intentions, is the fact that the intentional system feels (i.e., is conscious) whereas the systematically interpretable (and Turing-indistinguishable) syntactic system does not.* Ditto for intentional states:

The only difference between (a) a symbol system whose symbols are merely systematically interpretable as being about something (even in a robot that is Turing-indistinguishable from one of us) ("derived intentionality") and (b) a symbol system whose symbols are not only interpretable as being about something but really *are* about something ("intrinsic intentionality") is whether or not the symbol system feels.

One has at least two ways to deny this, but I don't think either of them will be very satisfying:

> (1) One can say that there is no difference between (a) a symbol system whose symbols are merely systematically interpretable as being about something and (b) a symbol system whose symbols really *are* about something. (But that would leave human minds indistinguishable from inert books or interactive online encyclopedias, or toy robots, or lifelong Turing-Test-Passing robots in the world; and that in turn would make "intentionality" a rather unremarkable (and nonmental) "mark" of the mental.)
>
> (2) If one does not want to say that there is no difference between (a) and (b), and one does not think that the difference is feeling, then one has to say (non-circularly, and non-emptily) what the difference is. (For me, there is no other substantive candidate in sight; indeed, the intentionalists don't even seem to realize that they need one!)

By way of an example, my unconscious "preference" for being complimented can even be demonstrated by Skinner, by rewarding me with a smile and an acquiescent nod every time I say the word "hence." The frequency with which I would preferentially use "hence" would increase, reliably, given this systematic reinforcement for a long enough time, yet I could and would say, truly, hand on heart, that I had been entirely unaware of Skinner smiling whenever I said "hence," nor was I aware that I was saying "hence" more often -- nor was I even aware that I preferred to be complimented!

This is, of course a standard form of implicit learning (Skinnerian, in this instance). One could also get this effect out of a relatively primitive robot that was merely wired to do more often whatever it does that is followed by reward. Again, there is no intentionality involved in *either* case, even though in the first (human) case (me) the system *does* have "intentionality" whereas in the second case it does not.

Nothing is really at stake in any of this for Darwin here, however.

---

**Galen Strawson** pointed out that William James (1890) wrote the following regarding the evolutionary origins of consciousness (cf. Strawson 2006):

> *"The demand for continuity has, over large tracts of science, proved itself to possess true prophetic power. We ought therefore ourselves sincerely to try every possible mode of conceiving the dawn of consciousness so that it may not appear equivalent to the irruption into the universe of a new nature, non-existent until then.*
>
> *"Merely to call the consciousness 'nascent' will not serve our turn. It is true that the word signifies not yet [p. 149] quite born, and so seems to form a sort of bridge between*

> *existence and nonentity. But that is a verbal quibble. The fact is that discontinuity comes in if a new nature comes in at all. The quantity of the latter is quite immaterial. The girl in 'Midshipman Easy' could not excuse the illegitimacy of her child by saying, 'it was a very small one.' And Consciousness, however small, is an illegitimate birth in any philosophy that starts without it, and yet professes to explain all facts by continuous evolution.*
>
> *"If evolution is to work smoothly, consciousness in some shape must have been present at the very origin of things. Accordingly we find that the more clear-sighted evolutionary philosophers are beginning to posit it there. Each atom of the nebula, they suppose, must have had an aboriginal atom of consciousness linked with it; and, just as the material atoms have formed bodies and brains by massing themselves together, so the mental atoms, by an analogous process of aggregation, have fused into those larger consciousnesses which we know in ourselves and suppose to exist in our fellow-animals. Some such doctrine of atomistic hylozoism as this is an indispensable part of a thorough-going philosophy of evolution. According to it there must be an infinite number of degrees of conscious- [p.150] ness, following the degrees of complication and aggregation of the primordial mind-dust. To prove the separate existence of these degrees of consciousness by indirect evidence, since direct intuition of them is not to be had, becomes therefore the first duty of psychological evolutionism."* [*Principles of Psychology* [p. 148]]

To my ears, to reply to the question

> "How and why (and when, and since when) do some combinations of matter (sometimes) feel?"

with

> "All 'matter,' at all scales, and in all combinations, feels, always"

sounds not only extremely ad hoc, but exceedingly implausible (if not incoherent), even when it comes from the pen of William James. (Nor does it even begin to address the real underlying problem, which is causality.)

I continue to use "feels" systematically in place of "is conscious" or "has a mind," not only because they are all completely co-extensive, but because "feels" wears the real problem frankly and tellingly (and anglo-saxonly) on its sleeve, whereas most of the other synonyms and euphemisms -- especially "intentionality" -- obscure and equivocate.

The mark of the mental is and always was *feeling*. Without feeling, all that's left is mindless "functing" -- which is all there is outside the biosphere (until/unless exobiology provides evidence to the contrary), or inside the atom, or inside any feelingless combination of matter, even if it does computations that are systematically interpretable as semantically meaningful -- and even if the symbols in its computational "language of thought" are robotically grounded in the world via transducers and effectors, and even at Turing-Test-scale -- *as long as it does not* *feel* (Harnad 2008b).

And of course the issue is about *whether* something is being felt at all, not about *what* is being felt, or *how much*. So "degrees of consciousness" are completely irrelevant: We are talking about an all-or-none, 0/1 phenomenon.

---

**Tom Nagel** noted that perhaps there is something like the "poverty of the simulus" problem for genetic traits, in that 4 billion years don't seem enough to converge on the current outcomes by chance alone (cf. Nagel 1999).

I am a complete outsider to such calculations, but several thoughts come to mind:

> (1) For any long enough random time-series, the probability that the current state is reached from the initial state is just about zero, so that cannot be the right way to reckon it.
>
> (2) In the case of life, the initial conditions matter too, because at first the alternatives were much tighter (and they began even before the gene).
>
> (3) In many ways, evolution is playing chess with itself, because the "environment" of the succeeding generation of the genotypes of organisms consists to a great extent of the genotypes of other organisms (rather than just external things like, say, the weather on the planet). So probably this too focuses the options on something less than all combinatory possibilities.
>
> (4) It has often been pointed out that as organisms' genotypes and survival/reproduction contingencies became more structured across evolutionary time (as a result of random mutations and selective retention), more and more of the adaptive variation becomes just (random) variation in the timing and recombination of existing traits, rather than direct random genetic mutations that create new structures per se. (This is where the "evo-devo" principle comes from.)
>
> (5) Aside from all that, in the one case where the notion of the "poverty-of-the-stimulus" has been made explicit enough to formulate without hand-waving, the case of Universal Grammar (UG), the evidence is as follows (and I don't think the case of heritable biological traits in general conforms to this very specific and special paradigm):
>
>> (5a) It turns out that all existing languages are compliant with Universal Grammar (UG). UG consists of the rules that determine which utterances are and are not grammatically well-formed.
>>
>> (5b) UG is not taught to us, and we do not know the rules of UG explicitly; but we know them "implicitly," because we are able to produce all and only UG-compliant utterances, and to perceive when utterances violate UG.

(5c) Not only is UG not taught to us -- indeed, its rules are not even all known yet, but are still being explicitly learned (sic) gradually and collaboratively by generations of linguists, through trial and error hypothesis testing -- but UG cannot be learned through trial and error by the child in its language-learning years, because the child does not have enough evidence or time to learn UG. (This is the "poverty-of-the-stimulus.")

(5d) "Poverty-of-the-stimulus" has a very specific meaning in the case of UG: The rules of UG *can* be learned by trial and error from data (they are indeed being gradually learned by the generations of linguists since Noam Chomsky first posited their existence in the mid-50s). But in order to learn which utterances are and are not UG-compliant (so as to find the rules that will produce all and only the UG-compliant ones), it is necessary -- as in learning to recognize what is and is not in any category -- to sample enough instances of UG-compliant and non-UG-compliant utterances to be able to infer the underlying rules that will successfully decide all further new cases (an infinite number of them). But the child hears and produces only UG-compliant utterances.

(5e) To have any chance of learning the rules of UG from the data, as the generations of linguists are doing, the child would need to hear or produce non-UG-compliant utterances, and be corrected, or at least be told that they are ungrammatical. This virtually never happens. (The child produces, and is corrected for, grammatical errors, but not UG errors, because hardly anyone ever makes a UG error -- except linguists, deliberately, in testing their hypotheses about what the rules of UG are.)

(5f) Hence, as the child cannot be learning them, the child must already be born knowing (implicitly) the rules of UG.

For this sort of poverty-of-the-stimulus problem to arise with the Darwinian evolution of heritable traits, it would have to be the case that there was *no prior maladaptive variation*: The adaptive traits occurred, but the maladaptive ones did not. But that is not the case with evolution at all: Maladaptive traits occur (by chance) all the time, and are then "corrected," by the fact that they either handicap or make survival/reproduction impossible.

So the evolutionary counterpart of the poverty-of-the-stimulus (for heritable traits in general) would not be that it simply looks too unlikely that today's traits arose out of random recombinations and selective survival/reproductive success over 4 billion years of trial-and-error-correction. The evolutionary counterpart of the poverty-of-the-stimulus would be that maladaptive traits never occurred (or did not occur often enough) to be "corrected" by natural selection.

There is, however, one prominent exception to this exemption of evolution from the poverty-of-the-stimulus problem, and that is the evolution of UG itself!

Not to put too fine a point on it, but it is not at all apparent how there could be an adaptive story about how the rules of UG, heritably encoded in our brains, could have evolved in the same way that hearts, lungs, wings or eyes evolved (the usual trial-and-error-correction story, guided by the adaptive advantages/disadvantages). Linguists only needed a few generations to learn (most of) UG by trial and error, to be sure. But there does not seem to be any plausible way to explain how (i) what it was that those linguists were actually doing during those generations -- when their deliberate errors and hypotheses were being "corrected" by consulting the grammatical intuitions about what is and is not UG-compliant that were already built into their brains (by evolution, presumably) -- can be translated into (ii) an adaptive scenario for what our ancient ancestors were doing at the advent of language, *when their "errors" werebeing corrected instead by the adaptive disadvantages of trying to speak non-UG-compliantly*!

So unless UG turns out to be homologous with (a free-rider on?) some other trait that *does* have a plausible adaptive history, it looks as if the poverty-of-the-stimulus problem afflicts not only the learnability of UG by the child, guided by external error-correction for non-UG-compliant utterances: it afflicts also the evolvability of UG, guided by the maladaptive consequences of non-UG-compliant utterances.

Chomsky's own view is that there is something about the very nature of (linguistic) thinking itself, such that only UG-compliant thought is possible at all. So whatever made (linguistic) thinking adaptive for our ancestors necessarily made UG-compliant verbal thinking adaptive (because non-UG-compliant thinking is simply impossible). (No one has yet shown, however, how/why non-UG-compliant thinking would be impossible.)

Fodor, I think, overgeneralizes this intuition of Chomsky's concerning thinking itself, suggesting that it is not only UG that is native to (the language of) thought, but the lexicon too, or at least the terms with "simple" rather than composite referents and meanings ("big" and "dog," perhaps, but not "big dog" or "standard poodle" -- to put this in the doubly contentious vocabulary of animal breeding and its products!). (This is perhaps a legacy of Fodor's having taken on "generative semantics" while Chomsky handled generative grammar, way back when; Fodor & Katz 1964.)

I think Fodor is wrong about the lexicon, simply because *there is no poverty-of-the-stimulus problem for the learning of word meanings* (nor for the learning of the sensorimotor categories that often precede them): We encounter plenty of dogs and non-dogs, and we get plenty of corrections for calling dogs non-dogs and vice versa. And this is true for many of the (content) words in our dictionaries, whether composite or simple (so that once we have learned enough of them directly, we can learn the rest from recombinatory definitions based on the words we already know; Blondin-Massé et al. 2008).

But in trying to extrapolate the poverty-of-the-stimulus problem to the lexicon -- arguing that just as learning was incapable of accounting for how we came to have UG, learning is likewise incapable of accounting for how we came to have the "concepts" underlying our words -- Fodor found himself up against the other potential explanation of the origin of concepts: evolution.

So (if my psychoanalysis is correct!), Fodor has been trying to argue that evolution is no better able to explain origins (whether the origins of UG, or of "concepts," or even of heritable traits) than learning is. (My own view is that for UG, Fodor is right about both learning and evolution; for (most) "concepts" he is already wrong about learning, so he doesn't even need to consider evolution, because it's unnecessary; and for heritable traits in general, he is wrong about evolution.)

Moreover, like Chomsky, Fodor too eventually puts the onus on thought, suggesting that "concepts" are neither learned nor evolved, but somehow "endogenous" to thinking itself or the capacity to do it.

It could be. It could be that not only UG but the lexicon are so part and parcel of the very possibility of thinking at all that there is no independent story to be told about their provenance in terms of either learning or evolution. They might both be as "endogenous" to the possibility of doing thinking as the eternal Platonic truths (about, say, prime numbers) are intrinsic to the possibility of doing mathematics. Maybe the lexicon just comes with the territory for anyone or anything that thinks in the language of thought (Fodor 1975). Who knows? I doubt it, but that may just be because I am lacking the right abductive intuition about this...